\documentclass[11pt]{article}

\usepackage[preprint]{acl}

\usepackage{times}
\usepackage{latexsym}

\usepackage[T1]{fontenc}
\usepackage[utf8]{inputenc}
\usepackage{microtype}
\usepackage{inconsolata}
\usepackage{graphicx}

\usepackage{amsmath}
\usepackage{booktabs}
\usepackage{tabularx}
\usepackage{multirow}
\usepackage[most]{tcolorbox}
\usepackage{fvextra}
\newcolumntype{Y}{>{\centering\arraybackslash}X}
\definecolor{skillcardborder}{RGB}{205,205,205}
\definecolor{skillcardbg}{RGB}{252,252,252}
\definecolor{skilltagbg}{RGB}{238,238,238}
\newcommand{\skillcard}[3]{%
  \par\noindent
  \begingroup
  \setlength{\fboxsep}{4pt}%
  \setlength{\fboxrule}{0.35pt}%
  \fcolorbox{skillcardborder}{skillcardbg}{%
    \begin{minipage}{0.94\columnwidth}
      \raggedright
      {\small\colorbox{skilltagbg}{\strut\hspace{2pt}\textbf{Rule #1}\hspace{2pt}}}\par
      \vspace{1.5pt}
      {\small\bfseries #2}\par
      \vspace{1pt}
      {\small\color{black!76}#3}
    \end{minipage}%
  }%
  \endgroup
  \par\vspace{3pt}
}

\title{SKILL-KD: Contrastive Skill Distillation for LLM Agents}
\author{Qiming Shi\textsuperscript{\rm 1},
  Yibo Dou\textsuperscript{\rm 3},
  Jiawen Zhu\textsuperscript{\rm 2},
  Yulong Tao\textsuperscript{\rm 4},\\
  {\bf Linbo Jin\textsuperscript{\rm 4},
  Zhaolu Kang\textsuperscript{\rm 3},
  Yunfan Zhou\textsuperscript{\rm 1},
  Di Weng\textsuperscript{\rm 2}\thanks{\protect\hypertarget{Hfootnote.1}{Corresponding author.}}}\\
  \textsuperscript{\rm 1}State Key Lab of CAD\&CG, Zhejiang University\\
  \textsuperscript{\rm 2}School of Software Technology, Zhejiang University\\
  \textsuperscript{\rm 3}School of Software and Microelectronics, Peking University\\
  \textsuperscript{\rm 4}Alibaba Group\\
  \texttt{\{qimingshi,dweng\}@zju.edu.cn}}

\begin{document}

\maketitle

\begin{abstract}
Skill-based prompting has become a practical mechanism for improving LLM agents, yet existing methods often treat skills as summaries of the agent's own experience or of successful demonstrations.
This creates a mismatch for weaker student agents. A failed trajectory may not reveal the missing knowledge or strategy, while a teacher trajectory may be too implicit to internalize.
We propose SKILL-KD, a contrastive skill distillation framework that treats skills as an explicit distillation medium between agents of different capabilities.
Given a student failure and the teacher trajectory on the same task, SKILL-KD distills their actionable discrepancy into a textual skill patch, evaluates it by re-running the student, and iteratively refines it when the student still fails.
To prevent skill drift from repeated local updates, SKILL-KD maintains trace-linked edit histories and performs Drift-Aware Skill Consolidation to decide whether each patch is added, merged, or skipped.
Across five agent benchmarks and two student settings, SKILL-KD consistently improves frozen student agents over fixed-model adaptation baselines.
\end{abstract}

\begin{figure}[!t]
\centering
\includegraphics[width=\columnwidth]{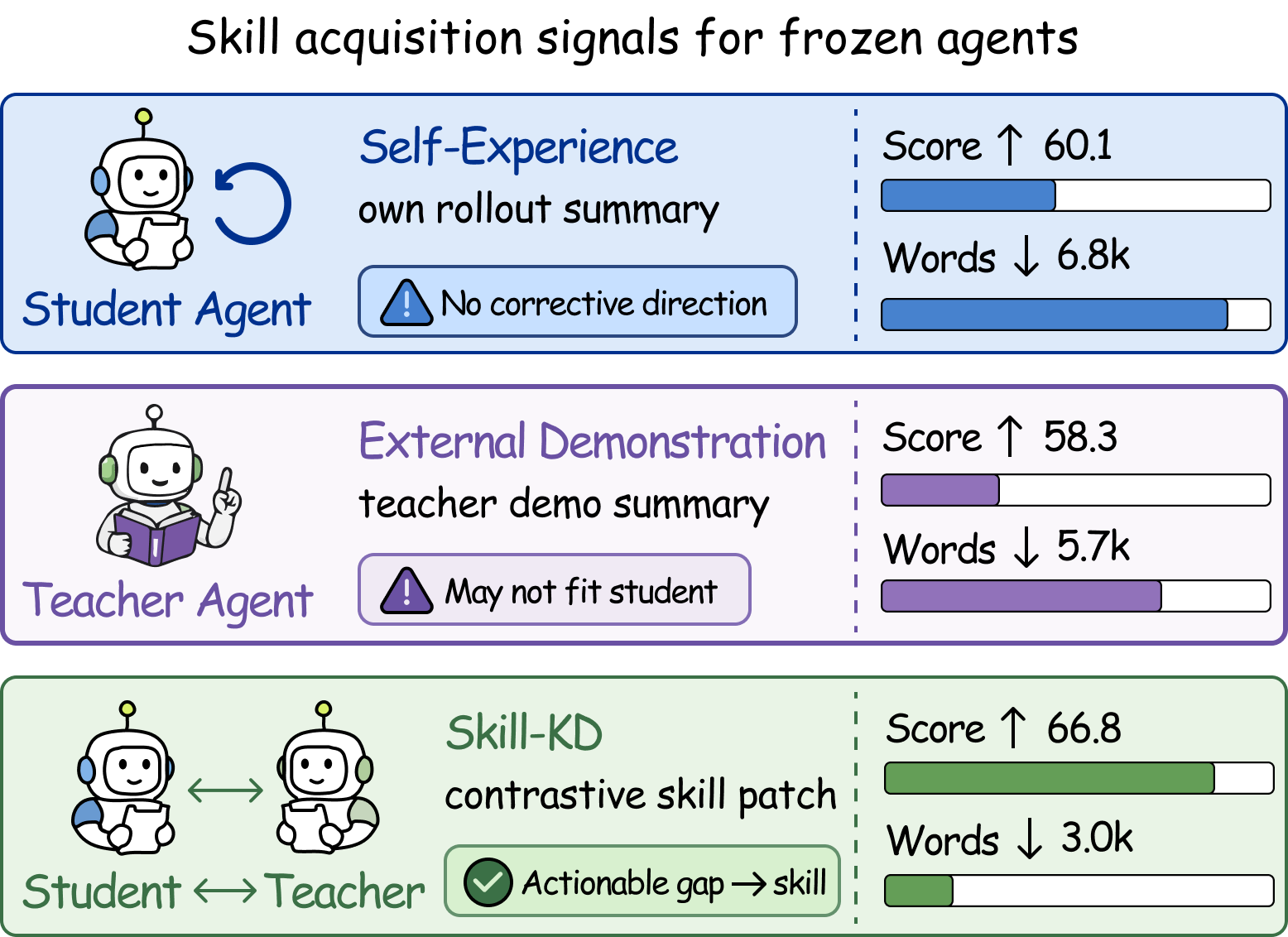}
\caption{Student-only skills may lack corrective guidance, while teacher-only skills may not fit the student.
SKILL-KD distills the teacher--student gap into actionable skills, yielding higher scores and a more compact skill library.}
\label{fig:intro-skill-signals}
\end{figure}

\section{Introduction}

Large language models are increasingly used as interactive agents that solve tasks through multi-step reasoning, tool use, and environment feedback.
In such settings, success depends not only on parametric knowledge, but also on reusable procedural patterns that guide how an agent explores an environment, invokes tools, decomposes a task, and verifies intermediate outcomes.
Recent work therefore studies skill-based agent systems, where successful trajectories, execution feedback, or reflections are distilled into explicit skills that can be retrieved and reused in later problem solving \cite{wang2023voyager,ni2026trace2skill,yang2026skillopt,ma2026skillgen,wang2026skillx}.

However, acquiring such skills remains challenging.
Existing approaches commonly derive reusable guidance either from the agent's own trajectories, reflections, or execution feedback \cite{shinn2023reflexion,zhao2023expel,ni2026trace2skill}, or from demonstrations, rationales, and trajectories provided by stronger models \cite{hsieh2023distilling,gu2024minillm,kang2025distilling,qiu2025agentdistill}.
In both cases, the difficulty is converting behavioral evidence into guidance the target agent can use. A failed rollout shows what went wrong but not what was missing, and a summary of a successful trajectory can be too abstract or misaligned to change future behavior.
As shown in Figure~\ref{fig:intro-skill-signals}, self-experience and external demonstrations provide helpful but indirect guidance, whereas SKILL-KD focuses on the actionable teacher-student gap.

We argue that skills should be viewed not merely as summaries of the agent's own experience or of successful demonstrations, but as a distillation medium between agents of different capabilities.
Classical knowledge distillation absorbs transferred knowledge into the student's weights~\citep{gou2021kdsurvey}, which is infeasible for closed-source students, opaque to inspection and editing, and requires retraining whenever the target tasks or environment change.
SKILL-KD therefore treats textual skills as the distillation carrier and keeps the student frozen.

Constructing this artifact requires identifying what the student should learn from the available behavioral evidence.
Rather than deriving guidance from the student's failed trajectory or transferring the teacher trajectory directly, SKILL-KD contrasts the two trajectories to identify the behavioral gap between them.
The key bottleneck is translating this gap into a textual skill patch that expresses the missing knowledge or strategy in a form the frozen student can follow.

SKILL-KD addresses this bottleneck through Adaptive Skill Distillation.
Instead of one-shot summarization, SKILL-KD searches the textual skill space iteratively, evaluating each candidate patch by observing how it changes the student's subsequent behavior.
If the student still fails, the consolidation agent refines the patch based on why the previous abstraction did not close the gap.
In this way, the method gradually compresses the teacher-student discrepancy into a student-usable skill, rather than assuming that a single teacher explanation is sufficient for transfer.

However, repeatedly producing local skill patches raises a repository-level challenge.
Skills are meant to encode how to solve a class of tasks, yet their update signals come from episodic evidence such as trajectories, reflections, or execution feedback.
This mismatch can cause \emph{skill drift}, where patches that fix current failures overfit to narrow cases, add redundant rules, or overwrite knowledge encoded by earlier skills.

SKILL-KD addresses this issue with Drift-Aware Skill Consolidation.
In addition to storing each patch, SKILL-KD maintains an edit log that records how the skill library has been modified over time and links every patch to its originating trajectories.
The consolidation agent therefore inspects the trace-linked edit history, which records why each patch was proposed, how related edits changed the student's behavior, and what happened to similar patches before.
Based on this evidence, the agent decides whether a patch adds a new skill, modifies or deletes an existing one, or is skipped as too case-specific.
This keeps the skill library compact and stable while reducing drift, without requiring all historical evidence to fit into the prompt at once.

Our framework provides a practical recipe for knowledge transfer among heterogeneous LLM agents.
It allows a strong teacher to improve a weaker student without weight updates, while maintaining a compact skill library grounded in observed teacher-student contrasts and validated through student behavior.
Our contributions are:

\begin{itemize}
    \item We formulate skill-level knowledge distillation for frozen LLM agents, where reusable textual skills serve as the distillation medium and teacher-student behavioral discrepancies provide the transferable learning signal.
    \item We propose SKILL-KD, a contrastive skill distillation framework that turns student-teacher trajectory pairs into validated skill patches through adaptive student reruns, and maintains a compact skill library through trace-linked drift-aware consolidation.
    \item We evaluate SKILL-KD across five agent benchmarks and two teacher-student settings, showing consistent gains over no-skill and skill-learning baselines, with ablations confirming the importance of adaptive validation and drift-aware consolidation.
\end{itemize}

\section{Related Work}

\subsection{Skill Acquisition and Optimization}

Skills have emerged as reusable procedural artifacts for adapting LLM agents without changing model weights~\cite{wang2023voyager,wang2026skillx,yang2026skillopt}.
One line of work distills skills from the trajectories of a stronger agent and transfers them to weaker agents as plug-and-play artifacts~\cite{wang2026skillx,qiu2025agentdistill,wang2026webxskill,fang2026memp}.
Another line evolves skills from the agent's own experience, either by distilling its trajectories into reusable rules~\cite{wang2023voyager,ni2026trace2skill,ma2026skillgen,zheng2025skillweaver} or by refining skills from execution feedback, akin to prompt optimization and experience-memory methods~\cite{yang2026skillopt,alzubi2026evoskill,wang2026skillgrad,yuksekgonul2024textgrad,pryzant2023protegi,agrawal2025gepa,shinn2023reflexion,zhao2023expel}.
In both cases skills are derived from one agent's experience, whereas SKILL-KD constructs them from same-task teacher-student behavioral contrast.

Beyond extraction, skill repositories must control redundancy, routing burden, and unsupported edits~\cite{yang2026skillopt,cho2026skillret,gao2026skillreducer,li2026graphskills,liu2026skillsvote}.
Existing consolidation relies on grouped trajectories or validation-set feedback, making its quality sensitive to trace-batch selection and validation coverage.
SKILL-KD removes this dependence by consolidating each patch against the trace-linked edit history of existing skills.

\begin{figure*}[t]
\centering
\includegraphics[width=\textwidth]{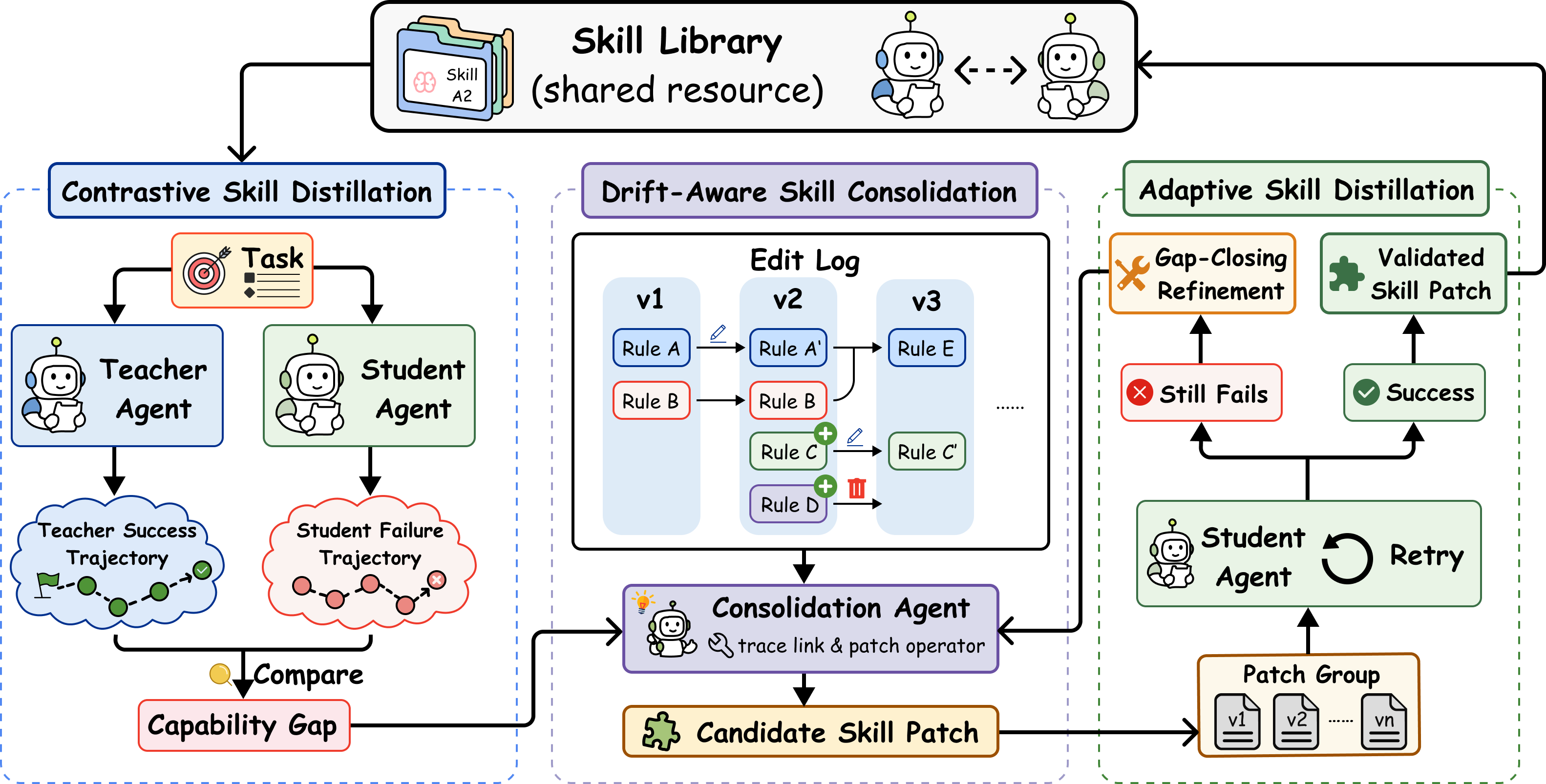}
\caption{Overview of SKILL-KD.
Given a task instance, the student and teacher generate contrasting trajectories.
The consolidation agent distills the teacher-student capability gap into candidate skill patches, adaptive reruns validate and refine those patches, and drift-aware consolidation updates a trace-linked skill library.}
\label{fig:method-overview}
\end{figure*}

\subsection{Knowledge Distillation for LLM Agents}

Knowledge distillation transfers stronger-model behavior to weaker models through soft targets, rationales, or trajectories~\cite{hsieh2023distilling,gu2024minillm}.
In agent settings, prior work distills tool-using behavior through trajectory-based training~\cite{kang2025distilling}, while AgentDistill distills reusable external modules from successful teacher trajectories without training~\cite{qiu2025agentdistill}.
Such guidance, however, may not be actionable for the weaker student.
Recent work instead uses skills or privileged signals as training-time supervision for policy or reinforcement learning~\cite{wang2026skillsd,yang2026opid,xia2026skillrl,yang2026rlsd}.
SKILL-KD differs by using same-task teacher-student trajectory contrast as the distillation signal and prompt-time textual skill patches as the transfer carrier, without updating the student's parameters.

\section{Method}

We present SKILL-KD, a skill-level distillation framework for transferring procedural knowledge from a stronger teacher agent to a weaker student agent without changing the student's model weights.
As shown in Figure~\ref{fig:method-overview}, the framework represents skills as trace-linked textual rules, then distills teacher-student behavioral discrepancies into candidate patches, adaptively validates them through student reruns, and finally consolidates locally useful edits into the global skill library.

\subsection{Contrastive Skill Distillation}

Let $\mathcal{D}_{\mathrm{train}}$ denote the training instances, $\mathcal{K}$ denote the current skill library, and $S$, $T$, and $C$ denote the student, teacher, and consolidation agent, respectively.
The skill library $\mathcal{K}$ is a structured Markdown document shared by the student and teacher.
Each skill in $\mathcal{K}$ is represented as
\[
p = (\mathrm{title}, \mathrm{content}, \mathrm{why}, \mathrm{trace}),
\]
where $\mathrm{title}$ names the patch, $\mathrm{content}$ specifies the actionable instruction, $\mathrm{why}$ records the rationale for the proposed update, and $\mathrm{trace}$ links it to the source trajectories.
The $\mathrm{title}$ and $\mathrm{content}$ fields are rendered into the shared skill file, while the $\mathrm{why}$ and $\mathrm{trace}$ fields are retained in an internal edit history accessible only to the consolidation agent.

Unlike parameter-level distillation, the student model is fixed.
The optimization target is not weights but the external skill library $\mathcal{K}$ that improves the frozen student's task performance across $\mathcal{D}_{\mathrm{train}}$.
For each training instance $x$, the student first attempts the task with the current skill library
\[
\tau_S^0 = S(x;\mathcal{K}).
\]
A task evaluator $r(x,\tau) \in [0,1]$ scores how well a trajectory $\tau$ solves instance $x$.
If $r(x,\tau_S^0)=1$, no update is needed and the framework proceeds to the next instance.
If the student fails, the teacher is run on the same instance
\[
\tau_T = T(x;\mathcal{K}).
\]
Even the teacher may not achieve a perfect score on every instance.
In such cases, comparing the two trajectories can still reveal where their intermediate decisions diverge and provide additional behavioral evidence for skill construction.
The consolidation agent combines this contrast with the trajectory-level evaluation scores to generate a candidate patch, whose utility is tested through the subsequent student rerun.

\subsection{Adaptive Skill Distillation}

The teacher-student gap provides a useful distillation signal, but identifying this gap alone is insufficient for reliable skill transfer.
Directly summarizing the contrastive difference into a rule does not guarantee that the rule will be effective under the student's policy.
The student may interpret the instruction differently or require a more specific formulation to change its behavior.
SKILL-KD therefore treats skill distillation as an adaptive search in textual skill space, where each candidate patch is validated by its actual effect on the student.

Given the student trajectory $\tau_S^0$, the teacher trajectory $\tau_T$, and the current skill library $\mathcal{K}$, the consolidation agent proposes an initial skill patch
\[
p^1 = C\bigl(\mathcal{K},\, \mathcal{H},\, (\tau_S^0, r_S^0),\, (\tau_T, r_T)\bigr),
\]
where $r_S^0 = r(x,\tau_S^0)$, $r_T = r(x,\tau_T)$, and $\mathcal{H}$ denotes the persistent edit history that records previous consolidation decisions and their associated trajectories.
We use $\mathrm{Apply}(p, \mathcal{K})$ to denote applying a patch $p$ to the skill library, which may add a new rule, modify an existing rule, or replace a previous formulation.
The student is rerun on the patched library
\[
\tau_S^1 = S\bigl(x;\, \mathrm{Apply}(p^1, \mathcal{K})\bigr).
\]
If $r(x,\tau_S^1)=1$, the patch is accepted into $\mathcal{K}$ and the process moves to the next instance.
If the student still fails, the consolidation agent observes the new student trajectory and proposes a revised patch.
We denote the accumulated evidence at step $i$ as $\mathcal{E}^i = \{(\tau_S^j, p^j, r_S^j)\}_{j=1}^{i}$.
The agent then produces
\[
p^{i+1} = C\bigl(\mathcal{K},\, \mathcal{H},\, (\tau_S^0, r_S^0),\, (\tau_T, r_T),\, \mathcal{E}^i\bigr),
\]
and the student is rerun with the latest patch $\tau_S^{i+1} = S(x;\, \mathrm{Apply}(p^{i+1}, \mathcal{K}))$.
Each $p^{i+1}$ is proposed against the original $\mathcal{K}$ and supersedes all previous patches for this instance.
This continues until the student succeeds or the maximum number of adaptive rounds $n$ is reached.
Only patches validated by successful student behavior are committed to the skill library.
If no adaptive round produces a successful trajectory within $n$ rounds, the instance produces no skill update.
Intermediate failed patches are not written to $\mathcal{K}$ or the persistent edit history $\mathcal{H}$ but are retained in the instance-level refinement record $\mathcal{E}^i$ during adaptive distillation.

\subsection{Drift-Aware Skill Consolidation}

The consolidation agent in the adaptive loop is not a single forward pass.
Making informed edit decisions over a growing skill library requires understanding why existing rules were introduced and how they relate to the current contrastive evidence.
We therefore implement the consolidation agent as a multi-turn agent that maintains a structured edit history and operates through explicit tool calls.

We define the edit history $\mathcal{H} = \{h_1, h_2, \ldots, h_m\}$ as a hierarchical record of all past modifications to $\mathcal{K}$.
Each entry $h_j$ stores
\[
h_j = (\mathrm{op}_j,\, p_j),
\]
where $\mathrm{op}_j \in \{\textit{add}, \textit{modify}, \textit{delete}, \textit{skip}\}$ is the operation performed, and $p_j = (\mathrm{title}, \mathrm{content}, \mathrm{why}, \mathrm{trace})$ is the associated patch as defined in the Constrastive Skill Distillation subsection.
The $\mathrm{why}$ field records the rationale for the operation, and the $\mathrm{trace}$ field indexes the originating trajectories that motivated this edit.
The trajectories themselves are not loaded into context by default, as they can be arbitrarily long, and are instead stored externally and retrieved on demand.

The consolidation agent operates through two tools.
The first is a \emph{trace-link} tool that, given an entry $h_j \in \mathcal{H}$, retrieves the full trajectories indexed by $\mathrm{trace}_j$.
The second is a \emph{patch} tool that commits an operation $\mathrm{op} \in \{\textit{add}, \textit{modify}, \textit{delete}, \textit{skip}\}$ to $\mathcal{K}$, inserting a new rule, updating or removing an existing rule, or leaving the library unchanged.
In practice, the consolidation agent invokes trace-link across multiple turns before calling the patch tool, retrieving trajectories behind related entries in $\mathcal{H}$ to identify cross-instance patterns, detect conflicts with existing skills, and determine if a new patch should be merged into an existing rule or added independently.
This multi-turn investigation keeps the skill library compact and stable while avoiding redundancy and gradual drift from repeated local edits.
The full prompt template, edit-log schema, and tool schemas are provided in Appendix~\ref{app:critic-prompt-tools}.

\section{Experiments}
\label{sec:experiments}

We evaluate whether contrastive skill distillation improves frozen student agents by transferring reusable knowledge from a stronger teacher, reporting each benchmark's native hard-success or exact-match score on the test partitions across SearchQA~\citep{dunn2017searchqa}, SpreadsheetBench~\citep{spreadsheetbench}, DocVQA~\citep{mathew2021docvqa}, LiveMath~\citep{he2026livemathematicianbenchlivebenchmarkmathematicianlevel}, and ALFWorld~\citep{alfworld}.

\begin{table*}[!t]
\centering
\small
\setlength{\tabcolsep}{3pt}
\renewcommand{\arraystretch}{1.08}
\begin{tabularx}{\textwidth}{@{}lYYYYYY@{}}
\toprule
\textbf{Method} & \textbf{SearchQA} & \textbf{Spreadsheet} & \textbf{DocVQA} & \textbf{LiveMath} & \textbf{ALFWorld} & \textbf{Avg} \\
\midrule
\multicolumn{7}{c}{\textbf{Base model: Qwen3.5-4B, teacher: Qwen3.7-plus}} \\
\addlinespace[1pt]
No Skill & 68.1 & 9.3 & 86.9 & 22.4 & 30.6 & 43.5 \\
EvoSkill & 68.6 & 17.9 & 79.7 & 23.4 & 67.9 & 51.5 \\
Trace2Skill & 68.5 & 19.3 & 88.0 & 27.2 & 64.9 & 53.6 \\
SkillGen & 69.2 & 13.2 & 88.8 & 21.0 & 70.1 & 52.5 \\
SkillOpt & 71.2 & 23.9 & 89.0 & 52.0 & 81.3 & 63.5 \\
\textbf{SKILL-KD} & \textbf{79.2} & \textbf{24.3} & \textbf{89.3} & \textbf{54.8} & \textbf{86.6} & \textbf{66.8} \\
\midrule
\multicolumn{7}{c}{\textbf{Base model: Qwen3.6-35B-A3B, teacher: ChatGPT-5.5}} \\
\addlinespace[1pt]
No Skill & 72.7 & 38.2 & 87.6 & 31.2 & 59.7 & 57.9 \\
EvoSkill & 75.6 & 32.9 & 89.8 & 32.3 & 79.1 & 61.9 \\
Trace2Skill & 75.4 & 33.2 & 90.4 & 29.6 & 70.9 & 59.9 \\
SkillGen & 75.2 & 40.0 & 90.1 & 36.3 & 63.4 & 61.0 \\
SkillOpt & 80.3 & 47.5 & 91.4 & 41.6 & 82.1 & 68.6 \\
\textbf{SKILL-KD} & \textbf{80.6} & \textbf{49.3} & \textbf{92.2} & \textbf{54.8} & \textbf{96.3} & \textbf{74.6} \\
\bottomrule
\end{tabularx}
\caption{Main results on held-out test splits.
Scores are percentages.
Avg is the macro-average over the five benchmark scores.
Bold marks the best score within each benchmark and block.}
\label{tab:main-results}
\end{table*}

\paragraph{Setting.}
We use two teacher-student configurations.
In Group~1, Qwen3.5-4B is the student and Qwen3.7-plus is the teacher; in Group~2, Qwen3.6-35B-A3B is the student and ChatGPT-5.5 is the teacher.
The consolidation agent uses the teacher model in each group, i.e., Qwen3.7-plus for Group~1 and ChatGPT-5.5 for Group~2.
For benchmarks with official train/validation/test partitions, we preserve the native splits.
When a benchmark does not define such partitions, we deterministically partition its evaluation pool using a $2{:}1{:}7$ train/validation/test ratio, the same default split protocol as SkillOpt~\cite{yang2026skillopt} (split seed $42$).
SKILL-KD constructs its skill library only from training instances; the validation portion is reserved for baselines whose original protocols require validation or selection, and is not used to train or select SKILL-KD skills.
Evaluation follows each benchmark's native hard-success or exact-match metric on the test partition.
Detailed benchmark-specific settings are in Appendix~A.

\paragraph{Baselines.}
We compare against four skill-acquisition baselines.
\emph{No Skill} is the original frozen student model without any external skills.
\emph{EvoSkill} iteratively analyzes execution failures, proposes new skills or edits to existing ones, and materializes accepted edits as structured skill folders~\citep{alzubi2026evoskill}.
\emph{Trace2Skill} consolidates broad execution trajectories into a unified skill directory through inductive reasoning over agent experience~\citep{ni2026trace2skill}.
\emph{SkillGen} synthesizes an auditable skill from base-agent trajectories, using successful and failed trajectories to induce reusable patterns and intervention-style verification to measure net skill effects~\citep{ma2026skillgen}.
\emph{SkillOpt} treats the skill document as an optimizable external state, applying bounded add/delete/replace edits from scored rollouts and accepting edits when held-out validation improves~\citep{yang2026skillopt}.

\subsection{Main Results}
\label{sec:main-results}

\paragraph{Overall Comparison.}
Table~\ref{tab:main-results} shows that SKILL-KD improves both student models across all five task families.
The evaluated tasks span search-based question answering, spreadsheet manipulation, multimodal document QA, mathematical reasoning, and embodied environment interaction.
For Qwen3.5-4B, the improvements are especially pronounced on SpreadsheetBench, LiveMath, and ALFWorld, while the method also yields gains on SearchQA and DocVQA.
For Qwen3.6-35B-A3B, whose No Skill baseline is already stronger, SKILL-KD still improves every benchmark and preserves clear gains on structured, mathematical, and embodied tasks.
This breadth suggests that the distilled skills capture reusable procedural guidance rather than benchmark-specific prompt tuning.

\paragraph{Homogeneous vs.\ Heterogeneous Distillation.}
Group~1 pairs Qwen3.5-4B with Qwen3.7-plus, a \emph{homogeneous} setting where both models belong to the same Qwen family.
Group~2 pairs Qwen3.6-35B-A3B with ChatGPT-5.5, a \emph{heterogeneous} setting where teacher and student come from different model families with distinct tokenizers, pretraining data, and behavioral characteristics.
SKILL-KD achieves consistent improvements in both regimes, confirming that contrastive skill distillation does not depend on architectural similarity.
Because the distillation carrier is a natural-language skill artifact rather than logits or hidden representations, the method remains agnostic to model internals and transfers across model families.

\paragraph{Generalization of Distilled Skills.}
ALFWorld's test split consists of unseen environments, providing a within-benchmark generalization check.
SKILL-KD improves Qwen3.5-4B from $30.6$ to $86.6$ and Qwen3.6-35B-A3B from $59.7$ to $96.3$ on this split, suggesting that the skill library captures reusable interaction strategies rather than memorizing training environments.

\section{Analysis}
\label{sec:analysis}

\subsection{Analysis of Skill Acquisition Signals}
\label{sec:ablation-distillation}

We compare four skill acquisition variants under the Group~1 setting.
\emph{Student-only} uses the student's failed trajectory for self-reflection.
\emph{Teacher-only} derives skills solely from the teacher's trajectory without observing the student's failure.
\emph{Batch} aggregates four trajectories into one skill update in a single pass.
\emph{Pairwise} performs a single contrastive skill update without adaptive refinement.

This comparison reveals two failure modes in skill construction without behavioral validation.
First, Teacher-only is the weakest learned variant ($58.3$, $+14.8$), showing that access to a stronger trajectory alone does not yield student-aligned guidance, as rules derived without the student's failure miss the misconception that caused it.
Second, Student-only reaches $60.1$ but accumulates $96$ rules and $6{,}821$ words, indicating that self-reflection produces local corrective heuristics rather than compact transferable rules.
Pairwise and Batch remain in between ($59.0$ and $59.4$) without resolving this tension.
SKILL-KD reaches $66.8$ with only $38$ rules.
Adaptive refinement is what converts teacher-student contrast into policy-aligned guidance by testing whether each patch actually changes student behavior.

\begin{table}[!htbp]
\centering
\small
\setlength{\tabcolsep}{1pt}
\renewcommand{\arraystretch}{1.05}
\begin{tabularx}{\columnwidth}{@{}lYYYY@{}}
\toprule
\textbf{Method} & \textbf{Avg. Score} & \textbf{Avg. Gain} & \begin{tabular}[c]{@{}c@{}}\textbf{Total}\\\textbf{Rules}\end{tabular} & \begin{tabular}[c]{@{}c@{}}\textbf{Total}\\\textbf{Words}\end{tabular} \\
\midrule
No Skill & 43.5 & +0.0 & 0 & 0 \\
Student-only & 60.1 & +16.6 & 96 & 6,821 \\
Teacher-only & 58.3 & +14.8 & 80 & 5,683 \\
Batch & 59.4 & +15.9 & 46 & 3,849 \\
Pairwise & 59.0 & +15.5 & 85 & 6,966 \\
\textbf{SKILL-KD} & \textbf{66.8} & \textbf{+23.4} & \textbf{38} & \textbf{3,010} \\
\bottomrule
\end{tabularx}
\caption{Effect of skill acquisition variants in Group~1 (Qwen3.5-4B student, Qwen3.7-plus teacher).
Avg.\ Score is the macro-average over the five benchmarks; Avg.\ Gain is measured against No Skill.
Total Rules and Total Words report the final skill-library size.}
\label{tab:ablation-distillation}
\end{table}

\subsection{Ablation Study}
\label{sec:ablation-study}

Table~\ref{tab:ablation-study} ablates the three core components of SKILL-KD under the Group~1 setting.
\emph{w/o Teacher} replaces the teacher trajectory with the student's own failed trajectory and lowers the average score from $66.8$ to $62.1$, showing that iterative refinement alone cannot substitute for the contrastive teacher signal.
\emph{w/o Adaptive} performs a single contrastive update without behavioral validation and reduces the average to $59.0$.
\emph{w/o Consolidation} removes edit-history access and hurts SpreadsheetBench and LiveMath most, where local edits accumulate incompatible rules.

The ablations suggest that the three components address complementary requirements for skill transfer.
Teacher-student contrast supplies corrective evidence, adaptive reruns test whether a proposed patch resolves the student's failure, and trace-linked consolidation uses historical evidence to integrate edits with existing skills.
The full framework achieves the highest average score, consistent with the contribution of all three components.

\begin{table}[!htbp]
\centering
\small
\setlength{\tabcolsep}{2pt}
\renewcommand{\arraystretch}{1.05}
\begin{tabularx}{\columnwidth}{@{}lYYYYYY@{}}
\toprule
\textbf{Variant} & \textbf{SQA} & \textbf{SS} & \textbf{Doc} & \textbf{LM} & \textbf{ALF} & \textbf{Avg} \\
\midrule
Full & \textbf{79.2} & \textbf{24.3} & \textbf{89.3} & \textbf{54.8} & \textbf{86.6} & \textbf{66.8} \\
w/o Teacher & 78.6 & 21.4 & 89.0 & 48.4 & 73.1 & 62.1 \\
w/o Adaptive & 76.4 & 17.5 & 88.0 & 27.4 & 85.8 & 59.0 \\
w/o Consolidation & 79.2 & 14.6 & 87.4 & 27.4 & 85.1 & 58.7 \\
\bottomrule
\end{tabularx}
\caption{Ablation of core components in Group~1 (Qwen3.5-4B student, Qwen3.7-plus teacher).
Scores are percentages on held-out test splits.
SQA, SS, Doc, LM, and ALF denote SearchQA, SpreadsheetBench, DocVQA, LiveMath, and ALFWorld.}
\label{tab:ablation-study}
\end{table}

\subsection{Effect of Trace-Linked Consolidation}
\label{sec:skill-library-compactness}

Table~\ref{tab:drift-library-stats} compares the final skill-library sizes with and without consolidation across all five benchmarks in Group~1.
Without edit-history access the final library grows to $85$ rules, with SpreadsheetBench alone accumulating $50$ rules ($2{,}930$ words), whereas trace-linked consolidation keeps the full library at $38$ rules ($3{,}010$ words) by merging new evidence into existing rules rather than appending new ones.
The more compact library also achieves a higher average test score ($66.8$ versus $58.7$; Table~\ref{tab:ablation-study}), indicating that compactness does not come at the expense of task performance.

\begin{table}[!htbp]
\centering
\small
\setlength{\tabcolsep}{3pt}
\renewcommand{\arraystretch}{1.05}
\begin{tabularx}{\columnwidth}{@{}lYYYYYY@{}}
\toprule
\textbf{Variant} & \textbf{SQA} & \textbf{SS} & \textbf{Doc} & \textbf{LM} & \textbf{ALF} & \textbf{All} \\
\midrule
w/o Consolidation & 17 & 50 & 9 & 5 & 4 & 85 \\
\textbf{w/ Consolidation} & \textbf{15} & \textbf{6} & \textbf{9} & \textbf{4} & \textbf{4} & \textbf{38} \\
\bottomrule
\end{tabularx}
\caption{Final skill-library size in rules per benchmark, comparing SKILL-KD without consolidation (w/o Consolidation) and the full SKILL-KD with trace-linked consolidation in Group~1 (Qwen3.5-4B student, Qwen3.7-plus teacher).
SQA, SS, Doc, LM, and ALF denote SearchQA, SpreadsheetBench, DocVQA, LiveMath, and ALFWorld.}
\label{tab:drift-library-stats}
\end{table}

\subsection{Qualitative Analysis of Skill Drift}
\label{sec:case-study}

Repeated local edits can gradually drift the skill library away from stable procedural knowledge.
To understand how such drift arises in practice, we inspect edit histories from variants with and without consolidation.
The cases reveal three recurring failure modes, namely case-specific drift, skill-bloat drift, and destructive-update drift.

\paragraph{Case-specific drift.}
Case-specific drift occurs when a skill that should serve as procedural long-term knowledge is repeatedly revised using only local evidence.
In DocVQA, the same numeric-extraction rule is edited over time as new cases arrive.
Its boundary first shifts toward preserving units for answers like \texttt{10 mg}, then toward dropping units already stated in questions such as \texttt{(kg)} or \texttt{\%}, and later shifts again for currency.
These edits are individually reasonable, but without trace-linked history the agent cannot see what evidence the rule already carries, so each patch can move the boundary toward the latest case.
The consolidation agent follows trace links to recover earlier cases and separates the evidence into stable rules, one omitting question-specified units or symbols and another preserving document-provided currency symbols.

\paragraph{Skill-bloat drift.}
Skill-bloat drift occurs when many validated patches remain correct in isolation but encode incidental task details.
In SpreadsheetBench, editing the skill  without edit history and trace-link access yields $50$ rules and $2{,}930$ words, including coordinate-level repairs tied to cells such as \texttt{P6:Q6}, \texttt{R3:R4}, and \texttt{C4:F6}.
These patches record real failures, but they increase routing burden and make the library less compact.
By retrieving related traces before patching, the consolidation agent merges repeated evidence into $6$ rules and $522$ words covering reusable operations such as dynamic lookup, tiered calculation, rolling-window aggregation, value replication, and numeric cleaning.

\paragraph{Destructive-update drift.}
Destructive-update drift occurs when iterative modifications overwrite previously useful skills.
In SpreadsheetBench, an earlier patch records the concrete tiered-bonus algorithm, including currency parsing and min/max clamping across income tiers.
A later conditional-sum patch pushes the same rule toward the broader instruction to write executable \texttt{openpyxl} code instead of pseudo-code.
The latter edit fixes a real failure, but it weakens the earlier calculation behavior.
The edit log and trace-link tool let the consolidation agent recover earlier trajectories before applying an update, so generic executable-code guidance is kept separate from the tiered-calculation rule rather than overwriting it.

\subsection{Sensitivity to Adaptive Rounds}
\label{sec:sensitivity}

We analyze how the number of adaptive rounds $n$ affects skill-training efficiency under the Group~1 setting (Qwen3.5-4B student with Qwen3.7-plus teacher).
For each round, we report the cumulative success rate on the training split, defined as the fraction of training instances whose student rollout succeeds by round~$i$ during skill construction.
We include $n{=}0$, the initial attempt before any rerollout is performed for the current instance.

Figure~\ref{fig:sensitivity-rounds} shows diminishing returns, while Table~\ref{tab:sensitivity-rounds-full} in Appendix~A gives the corresponding cumulative rates.
The first adaptive round is the dominant source of improvement, raising the average training success rate from $58.9\%$ to $67.7\%$ and accounting for most of the total gain achieved by $n{=}3$.
This indicates that rerunning the student with a candidate patch is an effective behavioral test, since many initially failed instances can be corrected once the skill is grounded in the student's actual response.
Later rounds provide smaller but still meaningful gains, improving the average to $70.8\%$ and then $72.7\%$.
These additional gains are concentrated in tasks such as SpreadsheetBench, LiveMath, and ALFWorld, where failures often involve multi-step tool use, mathematical option discrimination, or environment interaction and an initial patch may only partially close the behavioral gap.
The results therefore support a bounded adaptive procedure.
Most transferable corrections are discovered after the first validation round, while a small number of hard cases benefit from further refinement without requiring an unbounded search.

\begin{figure}[!htbp]
\centering
\includegraphics[width=\columnwidth]{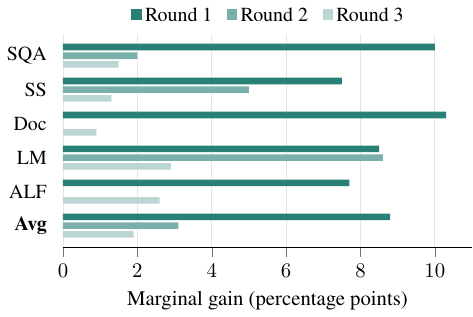}
\caption{Marginal percentage-point gains from successive adaptive rounds in Group~1 (Qwen3.5-4B student, Qwen3.7-plus teacher).
SQA, SS, Doc, LM, and ALF denote SearchQA, SpreadsheetBench, DocVQA, LiveMath, and ALFWorld.}
\label{fig:sensitivity-rounds}
\end{figure}

\subsection{Skill Distillation under Noise}
\label{sec:teacher-diagnostic}

Table~\ref{tab:teacher-diagnostic} reports teacher outcomes on training instances where the initial student rollout fails.
The teacher succeeds on $46.1\%$ of these cases on average, so skill construction does not assume oracle teacher supervision.
Patch acceptance is higher when the teacher succeeds ($45.3\%$) than when the teacher fails ($30.2\%$), showing that successful teacher trajectories provide stronger contrastive signals.
At the same time, teacher-failed cases still account for $39.2\%$ of accepted patches on average.
These results indicate that teacher failure does not preclude useful skill updates.
A failed teacher trajectory may still contain locally useful decisions or reveal an alternative intermediate strategy before failing elsewhere.
When contrasted with the student's failure, these partial differences can expose missing knowledge or ineffective behavior that is not apparent from either final outcome alone.
The consolidation agent abstracts this evidence into a candidate patch, while adaptive student reruns filter out patches that do not improve the student.

\begin{table}[!htbp]
\centering
\small
\setlength{\tabcolsep}{2pt}
\renewcommand{\arraystretch}{1.08}
\begin{tabularx}{\columnwidth}{@{}lYYYYYY@{}}
\toprule
\textbf{Metric} & \textbf{SQA} & \textbf{SS} & \textbf{Doc} & \textbf{LM} & \textbf{ALF} & \textbf{Avg} \\
\midrule
Teacher success & 44.4 & 37.3 & 61.1 & 10.0 & 77.8 & 46.1 \\
Accept $\mid$ T success & 56.8 & 40.0 & 72.7 & 0.0 & 57.1 & 45.3 \\
Accept $\mid$ T failure & 52.7 & 2.4 & 57.1 & 38.9 & 0.0 & 30.2 \\
Accepted from T failures & 53.7 & 9.1 & 33.3 & 100.0 & 0.0 & 39.2 \\
\bottomrule
\end{tabularx}
\caption{Teacher-outcome diagnostic on student-failed training instances in Group~1 (Qwen3.5-4B student, Qwen3.7-plus teacher).
Scores are percentages on the training instances.
Acceptance rows condition on teacher success or failure; Accepted from T failures is the fraction of accepted patches derived from teacher-failed rollouts (T denotes teacher).
SQA, SS, Doc, LM, and ALF denote SearchQA, SpreadsheetBench, DocVQA, LiveMath, and ALFWorld.}
\label{tab:teacher-diagnostic}
\end{table}

\section{Conclusion}

We introduced SKILL-KD, a contrastive skill distillation framework that improves frozen LLM agents by converting teacher-student behavioral discrepancies into reusable skills.
Instead of summarizing teacher demonstrations or student failures in isolation, SKILL-KD distills the actionable gap between them, validates each candidate patch through student reruns, and consolidates accepted patches using trace-linked edit histories.
Experiments across five benchmarks and two teacher-student settings show that the method consistently improves frozen students over no-skill and skill-learning baselines.
Ablations show that adaptive validation and drift-aware consolidation contribute to effective, compact skill libraries.
Together, these findings support textual skills as an editable interface for procedural knowledge transfer, with student behavior guiding what to retain.

\section*{Limitations}

Despite the effectiveness of SKILL-KD, our study has three main limitations.
First, SKILL-KD is evaluated on benchmarks with explicit success metrics, while open-ended agent tasks may require human, preference-based, or domain-specific feedback to validate candidate skills.
Second, the method depends on informative teacher-student behavioral contrasts, since the teacher need not be perfect, but the trajectories should reveal reusable corrections rather than identical failure modes.
Third, because the transferred knowledge is encoded as textual skills, the approach is most effective for procedural guidance that the frozen student can follow through prompting, and may need to be combined with retrieval, tool learning, or model adaptation for capabilities beyond natural-language instruction following.

\section*{Ethical Considerations}

This work adheres to the ACL Ethics Policy.
Skills generated by SKILL-KD may inherit limitations of the underlying models.
We recommend reviewing generated skills and maintaining human oversight in sensitive applications.
When sensitive data are involved, trajectory logs and derived skills should be checked and appropriately anonymized before sharing.

\bibliography{references}

\clearpage
\appendix
\section{Experimental Details}

\paragraph{Benchmark protocol and skill-library setup.}
We evaluate on five benchmarks, SearchQA, SpreadsheetBench, DocVQA, LiveMath, and ALFWorld.
For benchmarks with official train/validation/test partitions, we use the native partitions directly.
For benchmarks without an official split, we create deterministic train/validation/test partitions with a fixed split seed of $42$ and a default $2$/$1$/$7$ train/validation/test ratio.
SKILL-KD uses only the training partition for skill construction; the validation partition exists solely to run validation-dependent baselines under their published protocols.
The concrete split sizes are reported in Table~\ref{tab:appendix-splits}.
Teacher and student calls use medium reasoning effort by default.
SearchQA, DocVQA, and LiveMath use single-round question-answering or multiple-choice evaluation without an additional tool-call limit.
SpreadsheetBench uses multi-round spreadsheet-code execution with an \texttt{openpyxl} and \texttt{pandas} runtime and at most $12$ turns.
ALFWorld allows at most $50$ environment steps per episode.

\paragraph{Implementation and compute.}
The student models were served locally on a server equipped with four NVIDIA A100 80GB GPUs.
The teacher and consolidation models were accessed through their respective official APIs.
Each reported score was computed from a single complete evaluation run on the fixed test split.

\begin{table}[!htbp]
\centering
\small
\setlength{\tabcolsep}{2pt}
\renewcommand{\arraystretch}{1.08}
\begin{tabularx}{\columnwidth}{@{}lYYY@{}}
\toprule
\textbf{Dataset} & \textbf{Train} & \textbf{Validation} & \textbf{Test} \\
\midrule
ALFWorld & 39 & 18 & 134 \\
SpreadsheetBench & 80 & 40 & 280 \\
SearchQA & 400 & 200 & 1400 \\
LiveMath & 35 & 18 & 124 \\
DocVQA & 107 & 53 & 374 \\
\bottomrule
\end{tabularx}
\caption{Train, validation, and test split sizes used in our experiments.}
\label{tab:appendix-splits}
\end{table}

For each task instance, the maximum number of adaptive rounds is set to $n{=}3$ by default.
The cumulative training success rates underlying the adaptive-round sensitivity figure in the main text are reported in Table~\ref{tab:sensitivity-rounds-full}.

\begin{table}[!htbp]
\centering
\small
\setlength{\tabcolsep}{1pt}
\renewcommand{\arraystretch}{1.08}
\begin{tabularx}{\columnwidth}{@{}lYYYYYY@{}}
\toprule
\textbf{Round} & \textbf{SQA} & \textbf{SS} & \textbf{Doc} & \textbf{LM} & \textbf{ALF} & \textbf{Avg} \\
\midrule
$n{=}0$ & 75.3 & 16.3 & 83.2 & 42.9 & 76.9 & 58.9 \\
$n{=}1$ & 85.3 & 23.8 & 93.5 & 51.4 & 84.6 & 67.7 \\
$n{=}2$ & 87.3 & 28.8 & 93.5 & 60.0 & 84.6 & 70.8 \\
$n{=}3$ & 88.8 & 30.0 & 94.4 & 62.9 & 87.2 & 72.7 \\
\bottomrule
\end{tabularx}
\caption{Sensitivity to the number of adaptive rounds in Group~1.
Entries are cumulative training-set success rates (\%) by round~$n$.
Avg is the macro-average over the five benchmark scores.
SQA, SS, Doc, LM, and ALF denote SearchQA, SpreadsheetBench, DocVQA, LiveMath, and ALFWorld.}
\label{tab:sensitivity-rounds-full}
\end{table}

The student and teacher share the same benchmark-specific skill library $\mathcal{K}$ during skill construction.
Both agents see the rendered skill titles and contents, while rationale and trace metadata are retained only in the internal edit history used by the consolidation agent.
During training reruns and held-out evaluation, the agent receives the entire current skill file for that benchmark in its prompt.
We use this full-skill-file setting because this paper focuses on skill acquisition, adaptive validation, and consolidation rather than learned skill retrieval or routing.
We train and evaluate a separate skill library for each benchmark to keep comparisons controlled across benchmarks with different environments, tool interfaces, and evaluation metrics.

\paragraph{Baseline signal sources.}
Some baselines acquire skills without teacher trajectories, relying on the target agent's own rollouts, failures, reflections, or execution feedback.
Our ablations separate this signal-source difference from the effect of adaptive contrastive distillation.
Teacher-only tests whether teacher trajectories alone are sufficient by removing the student's failure signal, while Pairwise tests whether a single teacher-student contrastive update is sufficient by removing adaptive refinement.
Since both variants remain below full SKILL-KD, the observed gains are not explained by teacher access alone, but by validating and consolidating contrastive skill updates against the student's behavior.

\section{Consolidation Agent Prompt and Tools}
\label{app:critic-prompt-tools}

The consolidation agent is prompted as a skill curator.
It sees the current skill list, edit history, current task, student/teacher/retry trajectories, and the training evaluator record for each trajectory.

\begin{center}
\begin{tcolorbox}[
    breakable,
    colback=gray!5!white,
    colframe=gray!75!black,
    title=Consolidation Agent Prompt Template,
    arc=2mm,
    boxrule=0.5pt,
    left=1mm,
    right=1mm,
    top=1mm,
    bottom=1mm,
    width=\columnwidth
]
\small
\begin{Verbatim}[breaklines=true,breakanywhere=true,breaksymbolleft={},breaksymbolright={}]
System:
You are a skill curator. Maintain a compact list of
reusable rules an agent uses to solve tasks.

Each rule has three fields:
- title: a short noun phrase (<=8 words).
- content: one sentence in the form
  "When <trigger>, do <action>". The trigger must be
  type-level, not tied to filenames, cell addresses,
  room names, or specific object nouns. This is the
  only field shown to the student at runtime.
- why: one sentence explaining the failing rollout
  versus the succeeding rollout. This is audit-only.

You receive current skills and the full edit log.
Trajectories are ordered as student_initial,
teacher, then same-task retries. Each trajectory has
skill_used, model_output, and evaluation.
The edit log stores only rule-version metadata and
edit_index references; trajectory payloads are
retrieved by tool call when needed.

Decision discipline:
1. Classify existing rules as cover, partial, or
   unrelated.
2. Prefer update over add when a rule is partial.
3. Add only for a new reusable trigger.
4. Delete overlapping or low-value rules.
5. Avoid cosmetic rewrites and case-specific
   content.
6. End by calling submit_ops; use [] for no change.

User:
Current skill list:
{current_skill_list}
Edit log ({n_edits} prior edits, oldest first):
{edit_log}
New task:
{task_text}
Trajectories:
{json_trajectories}
Decide whether the skill list needs an operation.
\end{Verbatim}
\end{tcolorbox}
\end{center}

\begin{center}
\begin{tcolorbox}[
    breakable,
    colback=gray!5!white,
    colframe=gray!75!black,
    title=Edit Log Schema,
    arc=2mm,
    boxrule=0.5pt,
    left=1mm,
    right=1mm,
    top=1mm,
    bottom=1mm,
    width=\columnwidth
]
\small
\begin{Verbatim}[breaklines=true,breakanywhere=true,breaksymbolleft={},breaksymbolright={}]
Edit log ({n_edits} prior edits, oldest first):

[edit <edit_index>]
(group <group_index> round <round_index>)
<OP> rule <rule_id>
  before:       # present for update/delete
    title:   <previous rule title>
    content: <previous rule content>
    why:     <previous audit rationale>
  after:        # present for add/update
    title:   <new rule title>
    content: <new rule content>
    why:     <new audit rationale>

Example:
[edit 1] (group 0012 round 01) UPDATE rule 001
  before:
    title:   Preserve adjacent units
    content: When a source value has a unit,
             include the unit.
    why:     The student omitted 'mg'; the
             teacher preserved it.
  after:
    title:   Resolve answer units from question
    content: When the question names the unit,
             return only the value.
    why:     A retry showed unit preservation
             can over-specify answers.
\end{Verbatim}
\end{tcolorbox}
\end{center}

\begin{center}
\begin{tcolorbox}[
    breakable,
    colback=gray!5!white,
    colframe=gray!75!black,
    title=Tool Schema: get\_original\_trajectories,
    arc=2mm,
    boxrule=0.5pt,
    left=1mm,
    right=1mm,
    top=1mm,
    bottom=1mm,
    width=\columnwidth
]
\small
\begin{Verbatim}[breaklines=true,breakanywhere=true,breaksymbolleft={},breaksymbolright={}]
Input schema: get_original_trajectories
{
  "type": "function",
  "function": {
    "name": "get_original_trajectories",
    "description": "Fetch prior edit rollouts.",
    "parameters": {
      "type": "object",
      "properties": {
        "edit_index": {"type": "integer"}
      },
      "required": ["edit_index"]
    }
  }
}
Output schema: get_original_trajectories
{
  "type": "object",
  "properties": {
    "edit_index": {"type": "integer"},
    "group_index": {"type": "integer"},
    "round_index": {"type": "integer"},
    "op": {"type": "string"},
    "rule_id": {"type": "string"},
    "task_text": {"type": "string"},
    "student_trajectory": {
      "skill_used": "string",
      "model_output": "string",
      "evaluation": "object"
    },
    "teacher_trajectory": {
      "skill_used": "string",
      "model_output": "string",
      "evaluation": "object"
    }
  }
}
\end{Verbatim}
\end{tcolorbox}
\end{center}

\begin{center}
\begin{tcolorbox}[
    breakable,
    colback=gray!5!white,
    colframe=gray!75!black,
    title=Tool Schema: submit\_ops,
    arc=2mm,
    boxrule=0.5pt,
    left=1mm,
    right=1mm,
    top=1mm,
    bottom=1mm,
    width=\columnwidth
]
\small
\begin{Verbatim}[breaklines=true,breakanywhere=true,breaksymbolleft={},breaksymbolright={}]
Input schema: submit_ops
{
  "type": "function",
  "function": {
    "name": "submit_ops",
    "description": "Submit final skill edits.",
    "parameters": {
      "type": "object",
      "properties": {
        "ops": {
          "type": "array",
          "maxItems": 3,
          "items": {
            "type": "object",
            "properties": {
              "op": {
                "type": "string",
                "enum": ["add", "update", "delete"]
              },
              "id": {"type": "string"},
              "rule": {
                "type": "object",
                "properties": {
                  "title": {"type": "string"},
                  "content": {"type": "string"},
                  "why": {"type": "string"}
                }
              }
            },
            "required": ["op"]
          }
        }
      },
      "required": ["ops"]
    }
  }
}
Output schema: submit_ops
{
  "type": "null",
  "description": "Terminal tool."
}
\end{Verbatim}
\end{tcolorbox}
\end{center}

\section{Representative Final Skills}
\label{app:representative-skills}

The cards below list all final rules from the Group~1 4B skill library $\mathcal{K}$, grouped by benchmark.
To keep the appendix readable, long rendered contents are truncated after 28 words.
Following the representation in Section~3.1, each card reports only the fields rendered to the student and teacher; $\mathrm{why}$ and $\mathrm{trace}$ are retained only in the internal edit history.

\subsection*{SearchQA}

\skillcard{001}{Prefer categorical subtypes for type questions}{When a question asks for a 'type' or 'kind' of something and multiple candidates appear in context, prefer a distinct categorical subtype (a named variety) over a temporally ...}

\skillcard{002}{Parse 'this' question targets correctly}{When a question contains 'this' (either as 'this [noun]' or as a standalone pronoun) referring to an entity in a described relationship, identify what grammatical role that pronoun ...}

\skillcard{003}{Prefer primary entity over exhaustive lists}{When a question describes multiple instances, roles, or items but there is a single primary entity underlying them (either because the context emphasizes one as primary, or because ...}

\skillcard{004}{Preserve middle initials in person names}{When a person's full name appears in the context with a middle initial, include the middle initial in the answer rather than dropping it for conciseness --- but ...}

\skillcard{005}{Parse Jeopardy category-clue format correctly}{When context presents a Jeopardy-style entry in the format 'CATEGORY - Clue text', recognize that the text before the hyphen is the category name (not the answer), and ...}

\skillcard{006}{Avoid redundant context in answers}{When a question already mentions identifying context (brand name, location, category, or any qualifying term) or when an entity is commonly known by a shorter name, answer with ...}

\skillcard{007}{Match answer number to dominant context form}{When a question asks for an entity and the context predominantly uses the plural form of that entity (e.g., 'strawberries' appears multiple times), answer with the plural form ...}

\skillcard{008}{Answer entity-list questions with shared location or organization}{When a question consists only of a list of entity names without an explicit question word or verb, the question is asking what the entities have in common. ...}

\skillcard{009}{Answer location codes with location names}{When a question provides a location code or abbreviation (such as an airport code like PHL), answer with the full name of the location that the code represents, ...}

\skillcard{010}{Recognize quote and partial-phrase questions}{When a question is a partial phrase marked by ellipsis (e.g., '...on my mind') or a quote that is cut off mid-sentence (e.g., ending with a conjunction like ...}

\skillcard{011}{Match definition questions to precise term or number}{When a question provides a specific defining characteristic, description, or numerical claim (e.g., 'A house with floors that differ by about half a story' or 'accounts for a ...}

\skillcard{012}{Prefer original title over translation}{When context presents a work (film, book, etc.) with both an original foreign-language title and an English translation or release title, answer with the original title rather than ...}

\skillcard{013}{Resolve possessive pronouns in statement questions}{When a question is a declarative statement containing a possessive pronoun ('his,' 'her,' 'their'), the question is asking for the person the pronoun refers to --- answer with ...}

\skillcard{014}{Answer tribute band questions with original artist}{When a question names or describes a tribute band, cover band, or parody group, answer with the original artist, band, or work being tributed or covered, rather than ...}

\skillcard{015}{Answer adaptation questions with source material}{When a question names or describes a creative work (film, musical, book, etc.) and the context states that the work is an adaptation, re-telling, or version of another ...}

\subsection*{SpreadsheetBench}

\skillcard{001}{Time extraction handles datetime objects and strings}{When extracting a time component from a spreadsheet cell, check the type of the cell value.
If it is a datetime object, use the .time() method directly.
If ...}

\skillcard{002}{Spreadsheet list lookup with dynamic value reading}{When performing lookups against a list or range of values in the spreadsheet, read the lookup values dynamically from the spreadsheet cells rather than hardcoding them, identify columns ...}

\skillcard{003}{Tiered calculations with currency parsing}{When calculating tiered amounts based on value ranges, parse currency-formatted cell values (strings containing \$ and commas) to numeric values before calculations, use min/max bounds to compute amounts ...}

\skillcard{004}{Cell value replication writes values not formulas}{When a task requires replicating or referencing one cell's value in another cell, read the source cell's value and write it directly to the target cell (handling blank/None ...}

\skillcard{005}{Rolling workday window calculations}{When a task requires summing or averaging values over a rolling window of workdays, define workdays as Monday through Friday (weekday() < 5), start collecting workdays from the ...}

\skillcard{006}{Numeric character cleaning preserves only digits and decimals}{When a task requires removing non-numeric characters from cell values to produce clean numeric strings, keep only digits and decimal points (not commas), because commas in spreadsheet values ...}

\subsection*{DocVQA}

\skillcard{001}{Document field extraction}{When asked to find a specific field value in a structured document, scan for the exact field label and extract the text immediately adjacent to it, preserving the ...}

\skillcard{002}{Precise answer extraction from document subjects}{When extracting an answer from a document, provide only the specific entities directly requested, omitting supplementary context like related product names, usage details, descriptive labels (e.g., 'Page' when ...}

\skillcard{003}{Number formatting in extraction}{When extracting a number from a document where it appears with leading punctuation (like a hyphen or dash), remove the punctuation and provide only the numeric value.}

\skillcard{004}{Roman numeral to Arabic numeral conversion}{When extracting a diagram or figure number expressed as a Roman numeral, convert it to the corresponding Arabic numeral.}

\skillcard{005}{Handwritten digit disambiguation}{When extracting handwritten numbers, carefully examine ambiguous digits (such as 2 and 6) by checking for loop closures and tail shapes to avoid misrecognition.}

\skillcard{006}{Quotation mark normalization}{When extracting quoted terms or expressions from a document, normalize quotation marks to standard straight double quotes (") rather than preserving typographic curly quotes, as the expected answer ...}

\skillcard{007}{Printed text letter disambiguation}{When extracting text from a document, carefully verify each letter in proper nouns and company names, paying attention to easily confused letter pairs (such as 'a' and 'o') ...}

\skillcard{008}{Brand name extraction from logos}{When extracting a brand name from a logo or advertisement where the text appears with visual spacing but represents a single compound brand name, extract it as one ...}

\skillcard{009}{Currency symbol inclusion in monetary extraction}{When extracting a monetary amount from a document where a currency symbol (like \$) is present adjacent to the number, include the currency symbol directly attached to the ...}

\subsection*{LiveMath}

\skillcard{001}{PDE blow-up integral power selection}{When determining the strongest provable blow-up criterion for nonlinear PDEs, do not default to the L1-in-time gradient norm; verify if the specific energy estimates or equation structure ...}

\skillcard{002}{Selection of stronger result options}{When a multiple-choice question asks for the 'strongest statement' or 'best provable result' and offers a meta-option claiming a stronger result exists, select the meta-option, ...}

\skillcard{003}{Harmonic Analysis critical variation unboundedness}{When evaluating the \(L^p\) boundedness of \(r\)-variation norms for circular averaging operators in \(R^2\) with \(r=2\), do not assume boundedness for \(p>2\) based on maximal function results ...}

\skillcard{004}{Block-stable graph tree criticality}{When analyzing the block tree distribution of random graphs from a block-stable class relative to the critical threshold \(u_C\), recognize that the tree is subcritical for \(u<u_C\) ...}

\subsection*{ALFWorld}

\skillcard{001}{Examine object in lamp light}{When the task is to look at an object in light, pick up the target object first, then navigate to the lamp and use it.}

\skillcard{002}{Pick clean then place in receptacle}{When the task is to wash an object and put it away in a receptacle, first search likely locations (fridge, countertop, diningtable) to find and pick up the ...}

\skillcard{003}{Pick cool then place in receptacle}{When the task is to chill an object and place it on a receptacle, first search likely locations (countertops, stoveburners) to find and pick up the target object, ...}

\skillcard{004}{Pick heat then place in receptacle}{When the task is to heat up an object and place it in a receptacle, first search likely locations (countertops, diningtables, fridge) to find and pick up the ...}

\end{document}